\documentclass{article}

\PassOptionsToPackage{numbers, compress}{natbib}

\usepackage[preprint]{neurips_2020}

\usepackage[utf8]{inputenc} 
\usepackage[T1]{fontenc}    
\usepackage{hyperref}       
\usepackage{booktabs}       
\usepackage{amsfonts}       
\usepackage{nicefrac}       
\usepackage{microtype}      
\usepackage{todonotes}
\usepackage{graphicx}
\usepackage{caption}
\usepackage{subcaption}
\usepackage{color}
\usepackage{amsmath}
\usepackage{multirow}

\usepackage{algpseudocode,algorithm,algorithmicx}

\makeatletter
\newcommand{\algmargin}{\the\ALG@thistlm}
\makeatother
\newlength{\whilewidth}
\algdef{SE}[parWHILE]{parWhile}{EndparWhile}[1]
  {\parbox[t]{\dimexpr\linewidth-\algmargin}{%
     \hangindent\whilewidth\strut\algorithmicwhile\ #1\ \algorithmicdo\strut}}{\algorithmicend\ \algorithmicwhile}%
\algnewcommand{\parState}[1]{\State%
  \parbox[t]{\dimexpr\linewidth-\algmargin}{\strut #1\strut}}

\title{Low-Variance Policy Gradient Estimation with World Models}

\author{%
   Michal Nauman\\
   Universiteit van Amsterdam\\
   \texttt{nauman.mic@gmail.com} \\
   \And
   Floris den Hengst \\
   Vrije Universiteit Amsterdam \\
   \texttt{f.den.hengst@vu.nl} \\
}

\begin{document}

\maketitle

\begin{abstract}
    In this paper, we propose World Model Policy Gradient (WMPG), an approach to reduce the variance of policy gradient estimates using learned world models (WM's).
    In WMPG, a WM is trained online and used to imagine trajectories. The imagined trajectories are used in two ways. Firstly, to calculate a without-replacement estimator of the policy gradient. Secondly, the return of the imagined trajectories is used as an informed baseline.
    We compare the proposed approach with AC and MAC on a set of environments of increasing complexity (CartPole, LunarLander and Pong) and find that WMPG has better sample efficiency. 
    Based on these results, we conclude that WMPG can yield increased sample efficiency in cases where a robust latent representation of the environment can be learned. 
\end{abstract}

\section{Introduction}

Deep reinforcement learning (DRL) has shown impressive results in learning complex behaviours in high-dimensional high-entropy environments (\cite{mnih2015human}; \cite{silver2017mastering}; \cite{vinyals2019grandmaster}). DRL approaches are characterized by the use of neural networks to represent policy, value or other components of the MDP solution. While the end results are impressive, they are often achieved by training that takes days to complete. As such, sample efficiency is an important concern for DRL research. 

One of the approaches taken by the community is labeled 'world models' (WMs). In this approach, the agent learns a policy-agnostic representation of the environment, often built by the state, transition and reward networks (\cite{ha2018recurrent}; \cite{hafner2019dream}; \cite{kaiser2019model}; \cite{gelada2019deepmdp}). WMs were used to achieve control through planning (\cite{hafner2019learning}; \cite{van2020plannable}), but also gradient-based policy search, where policy rollouts are imagined by the WM (\cite{kaiser2019model}; and \cite{hafner2019dream}). Besides sample-efficient learning, WMs were shown to facilitate generalization \cite{kipf2019contrastive} and exploration \cite{sekar2020planning}. 

Alternatively, better sample efficiency can be achieved by building desirable properties of the estimators used for learning. In the context of the gradient-based policy search, variance of the gradient approximator can be reduced either by advantage \cite{schulman2015high} or by increasing the amount of samples, often with parallelization (\cite{mnih2016asynchronous}; \cite{asadi2017mean}; \cite{kool2019buy}). However, the variance reduction as a result of increasing the amount of samples in Monte Carlo (MC) approximation is rapidly diminishing. This is especially true for low entropy policies, where drawing additional independent samples might yield redundant results. To this end, (\cite{schulman2015trust}; \cite{kool2019buy}) consider approximating the policy gradient by sampling without replacement. The experiments shown promising results, but the requirement of the agent executing multiple trajectories until termination from many states is not scalable to real-world applications.

In this paper, we propose WMPG - a method for approximating low-variance policy gradient. WMPG uses environment interactions to learn a WM in an online fashion. The learned WM allows the policy gradient to be estimated without replacement, with Q-values calculated by finite-horizon forward-looking TD($\lambda$). Two approximation techniques yield a policy gradient with a favourable bias/variance trade-off, while the use of WM makes the method scalable to real-world problems. We compare the proposed method to AC and MAC on three environments of increasing complexity: CartPole; LunarLander and Pong. We show that the proposed method can achieve better sample efficiency as compared to the benchmark algorithms.

\begin{figure}[h!]
    \centering 
\begin{subfigure}{0.25\textwidth}
  \includegraphics[width=\linewidth]{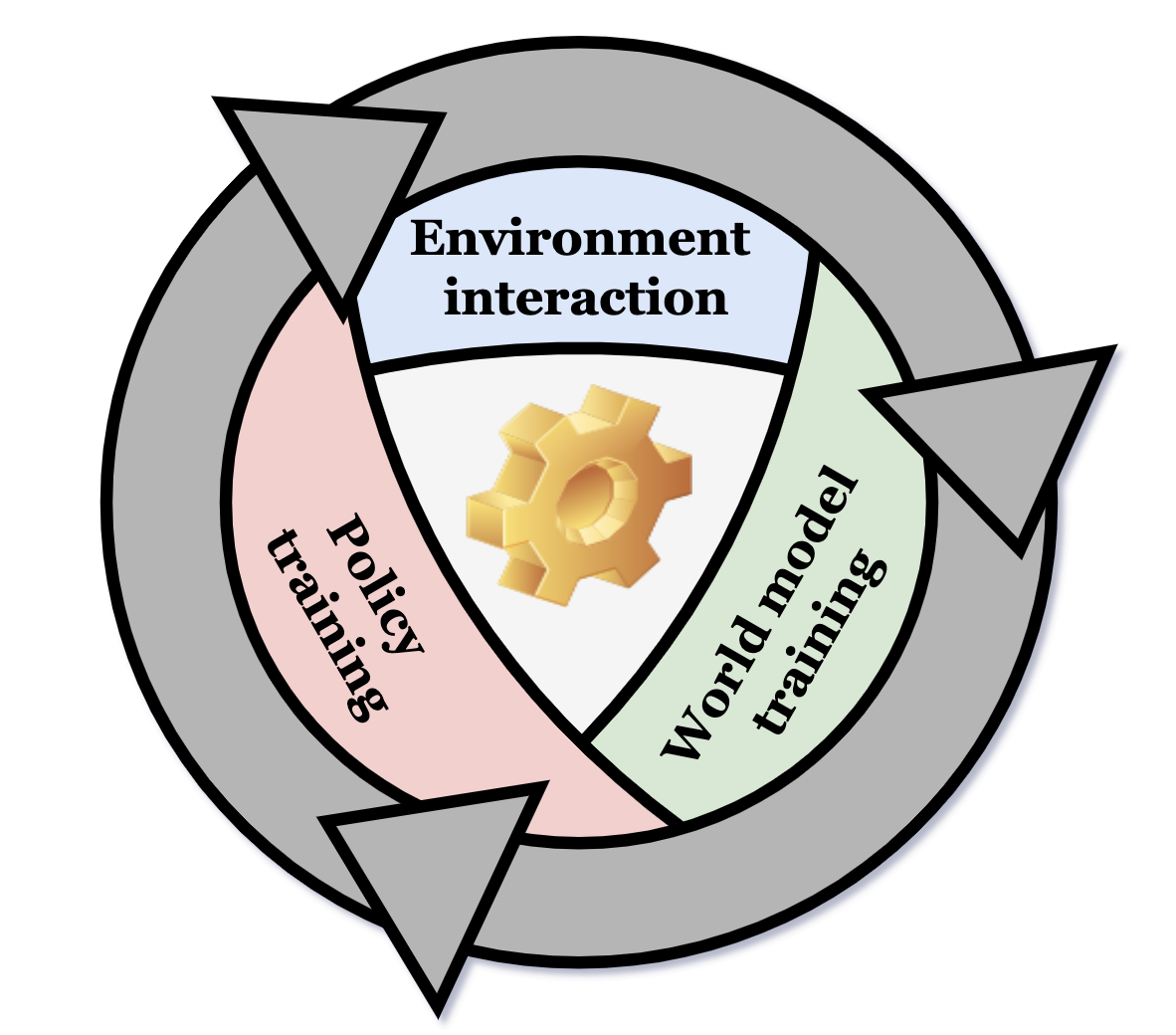}
  \caption{}
  \label{fig:o1}
\end{subfigure}
\begin{subfigure}{0.24\textwidth}
  \includegraphics[width=0.92\linewidth]{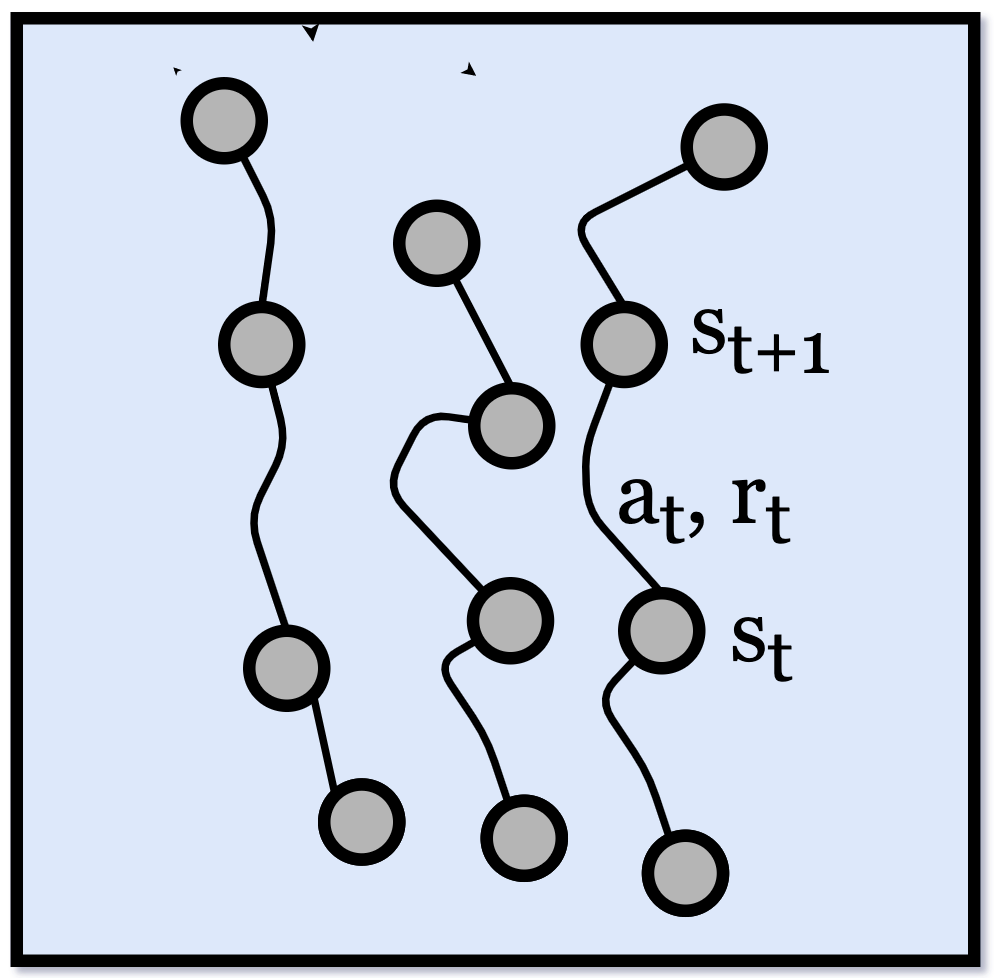}
  \caption{}
  \label{fig:o2}
\end{subfigure}
\begin{subfigure}{0.24\textwidth}
  \includegraphics[width=0.92\linewidth]{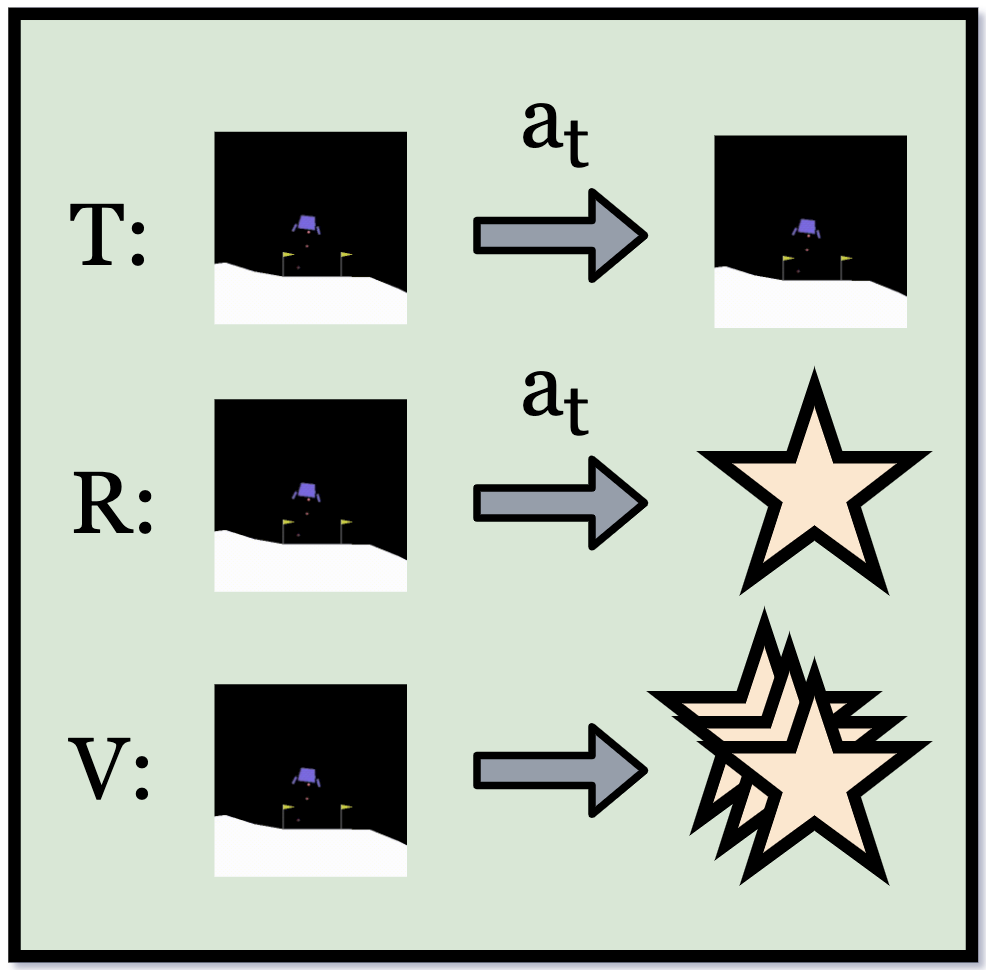}
  \caption{}
  \label{fig:o3}
\end{subfigure}
\begin{subfigure}{0.24\textwidth}
  \includegraphics[width=0.92\linewidth]{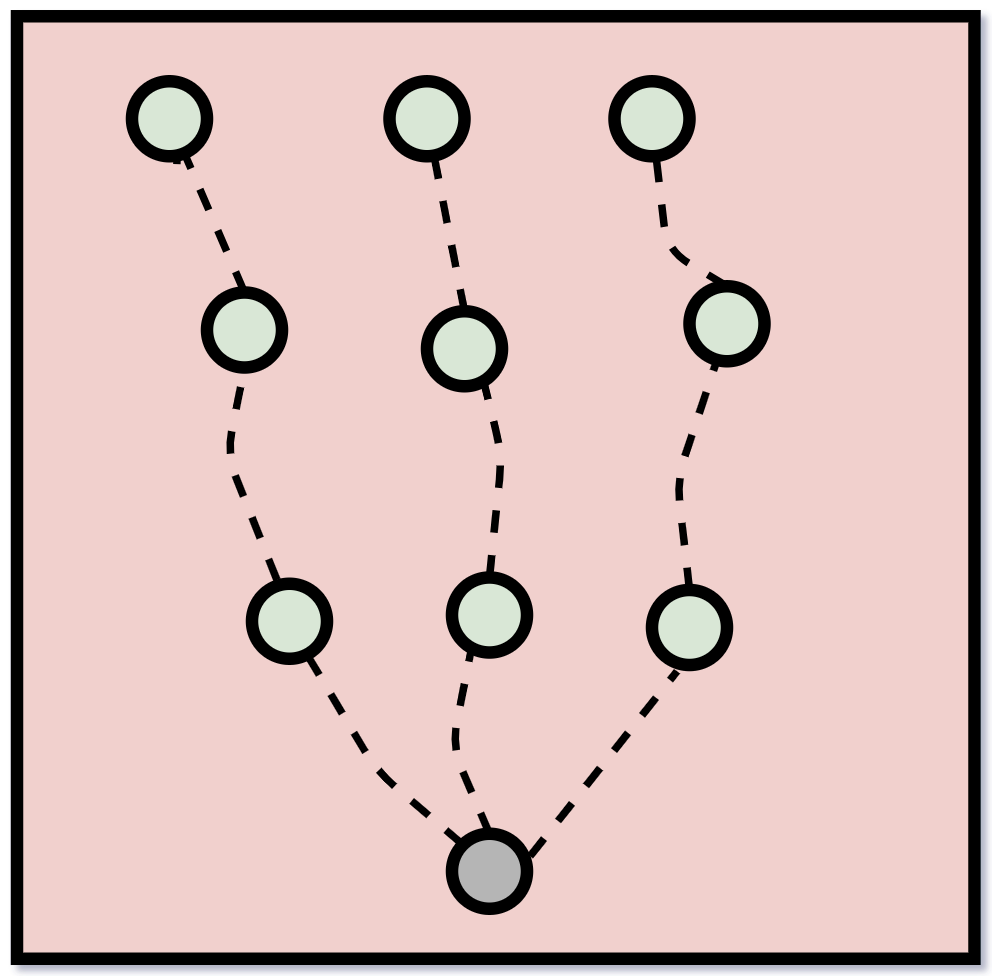}
  \caption{}
  \label{fig:o4}
\end{subfigure}
\caption{WMPG learning cycle. a) Agent gathers experience to train the WM that is used to estimate the policy gradient. b) Agent samples trajectories from the environment. The learning is triggered once enough experiences are gathered. c) Past transitions are used to train the transition and reward networks. d) For every state in the experience batch, agent samples multiple actions without replacement and imagines the Q-values of those actions.}
\label{fig:comp1}
\end{figure}

The main contribution of this paper is unifying no-replacement policy gradient approximation with world model framework. To the best of our knowledge, we are the first to propose: 1) using world models to calculate the policy gradient with a without-replacement expectation estimator; 2) using variable number of without-replacement sampled actions per state. We make our code available at: \url{https://github.com/WMPG-paper/WMPG}.

The paper is organized as follows. We firstly provide background for the problem that this paper addresses (policy gradient approximation) and the method that this paper uses (WMs). Next, we introduce the proposed approach, by listing the components of WMPG and explaining how WMPG calculates the policy gradient approximator. Finally, we discuss related work and the experiments performed for this paper.

\label{sec:intro}

\section{Background}
\label{sec:background}
\paragraph{Reinforcement Learning}
RL considers problems in the framework of Markov decision processes (MDPs). We consider deterministic MDPs with finite action sets. Thus, MDP is a tuple $\mathcal{M}:\langle \mathcal{S}, \mathcal{A}, \mathcal{T}, \mathcal{R}, \gamma \rangle$ where $\mathcal{S}$ is a continuous state representation, $\mathcal{A}$ is a finite set of actions with size $|\mathcal{A}|$, $\mathcal{T}: \mathcal{S} \times \mathcal{A} \to \mathcal{S}$ is a deterministic transition function, $\mathcal{R}: \mathcal{S} \times \mathcal{A} \to \mathbb{R}$ is a reward function and $\gamma \in [0,1]$ is a discount factor to balance current and future rewards. After observing state $s_t$ at each time step $t$, the agent performs action $a_t$ up to some end time $T$, following policy $\pi \in \Pi: \mathcal{S} \times \mathcal{A} \to [0,1]$ and collects reward $r_t = \mathcal{R}(s_t, a_t)$ at every time-step. The goal is to find a policy that maximizes the sum of collected rewards over time. 

\paragraph{Policy Gradient Methods} 

Gradient-based policy search defines a policy function $\pi_{\theta}$ that is differentiable wrt. to its parameters $\theta$ \cite{williams1992simple}. Then, $\theta$ is optimized such that the expected value of starting states is maximized, with expectation taken wrt. to the starting states. Assuming a single starting state, the gradient of the objective can be expressed as (\cite{sutton2000policy}; \cite{mnih2016asynchronous}):

\begin{equation}
    \nabla_{\theta}J(\theta) = \mathbb{E}_{s \sim \pi_{\theta}} \left[ \nabla_{\theta} J(\theta, s) \right] = \mathbb{E}_{s \sim \pi_{\theta}} \left[ \sum_{a \in A} Q^{\pi_{\theta}}(s, a) \nabla_{\theta} \pi_{\theta} (a|s) \right]
\end{equation}

By applying the log-derivative trick, the inner sum can be rewritten as expectation, such that policy gradient at state $s^{*}$ becomes equal to:

\begin{equation}
    \nabla_{\theta}J(\theta, s^{*}) = \mathbb{E}_{a \sim \pi_{\theta}} \left[ Q^{\pi_{\theta}}(s^{*},a) \nabla_{\theta} \log \pi_{\theta} (a|s^{*}) \right]
\end{equation}

In most practical applications, agent can perform only one action before being transitioned to some further state. As a result, the above expectation is often evaluated with single-sample MC, such that $\nabla_{\theta}\hat{J}(\theta, s^{*}) = Q^{\pi_{\theta}}(s^{*},a) \nabla_{\theta} \log \pi_{\theta} (a|s^{*})$ with $a \sim \pi_{\theta}$ (\cite{schulman2015high}; \cite{mnih2016asynchronous}; \cite{kool2019buy}). Given single-sample estimation with some action $a_i$, the sign of $\hat{J}(\theta, s^{*})$ depends solely on the sign of $Q^{\pi_{\theta}}(s^{*},a_i)$. 

Furthermore, $Q^{\pi_{\theta}}(s^{*}, a)$ might itself be unknown. If so, $\hat{Q}^{\pi_{\theta}}(s^{*}, a)$ can be calculated simultaneously with $\nabla_{\theta}\hat{J}(\theta, s^{*})$ using value approximation techniques like MC policy rollout, TD($n$) or TD($\lambda$) (\cite{schulman2015high}; \cite{feinberg2018model}; \cite{hafner2019dream}). Given some method of Q-value approximation, the non-zero variance of $\nabla_{\theta}\hat{J}(\theta, s^{*})$ can be reduced with an additive control variate $b(s)$. Then, the term $(\hat{Q}^{\pi_{\theta}}(s^{*},a) - b(s))$ is referred to as advantage. Having calculated $\nabla_{\theta}\hat{J}(\theta, s)$ for every state in batch $D_s$, the scalar loss is calculated with the batch average $\nabla_{\theta}\hat{J}(\theta) = \frac{1}{|D_s|} \sum_{s \in D_s} \nabla_{\theta}\hat{J}(\theta, s)$.

\paragraph{World Models} The framework of world models is designed to achieve optimal control with learned components of the underlying MDP. WMs are often represented by state embedding, transition and reward networks (\cite{ha2018recurrent}; \cite{kaiser2019model}; \cite{hafner2019dream}; \cite{van2020plannable}). Since successful control largely depends on the problem representation, a lot of attention is focused on learning latent representations that facilitate policy search (\cite{gelada2019deepmdp}; \cite{kipf2019contrastive}). Particularly, two classes of methods seem to emerge from the recent literature: reconstruction (\cite{ha2018recurrent}; \cite{igl2018deep}; \cite{hafner2019learning}; \cite{hafner2019dream}; \cite{kaiser2019model}); and bisimulation approximation (\cite{kipf2019contrastive}; \cite{gelada2019deepmdp};  \cite{van2020plannable}).

In reconstruction, original state representation is auto-encoded and the representation loss is placed in the original problem space. Thus, the state embedding is trained independently of the policy loss and other WM components. This method is known to have failure modes, especially when defined in the pixel space. Latent representations learned through reconstruction were shown to ignore important objects that are visually small (ie. ball in Pong) or use model capacity on rich backgrounds, which might be irrelevant from the perspective of state value (\cite{kaiser2019model}; \cite{kipf2019contrastive}). 

Bisimulation is guaranteed to be value preserving (\cite{larsen1991bisimulation}; \cite{li2006towards}).  Bisimulation can be approximated with bisimulation loss \cite{gelada2019deepmdp}. As the objective is placed outside of the pixel space, the approach directly tackles reconstruction's modes of failure. However, it was shown to map all states to a single point for sparse reward environments \cite{kipf2019contrastive}. The proposed solution, contrastive bisimulation \cite{van2020plannable}, was not tested in the context of on-policy WM learning.

Learning robust latent state representations is orthogonal to this paper. Recent research has shown approaches of learning latent representation for RL (\cite{ha2018recurrent}; \cite{hafner2019learning}; \cite{hafner2019learning}; \cite{kaiser2019model}; \cite{zhang2019solar}; \cite{baradel2020cophy}; \cite{biza2020learning}). Contrary to that work, this paper shows how learned latent models can be used to approximate low-variance policy gradient.

\section{World Model Policy Gradient}
\label{sec:wmpg}
WMPG learns a world model and uses it to approximate $\nabla_{\theta}J(\theta, s)$ for every state in the data batch. To calculate $\nabla_{\theta}J(\theta, s)$, WMPG samples multiple actions without replacement and imagines trajectories of length $h$ starting with those actions. Q-values of those actions are calculated using TD($\lambda$) calculated over the imagined trajectories. Finally, the without-replacement value estimator is used as a baseline. WMPG uses two components with independent memory buffers: a WM with a state embedding, a transition and a reward networks; and a behavior model consisting of a policy and a value networks.


\paragraph{State embedding} $Z_{\mu}: \mathcal{S} \rightarrow \mathcal{Z} \in \mathbb{R}^m$ maps the original state representation into some smaller dimensionality of size $m$. State dimensionality reduction allows the agent to learn more robust reward and transition approximators, as well as reduce the complexity of the policy search. Learning optimal state representation in the context of RL is an open research topic. As such, we learn the state compression with variational reconstruction in the pixel space, which was found to be working relatively well in multiple tasks.

\paragraph{Transition network} $T_{\kappa}: \mathcal{Z} \times \mathcal{A} \rightarrow \mathcal{Z}' \in \mathbb{R}^m$ approximates the transitions of the underlying MDP. Given that the underlying MDP is deterministic, transition can be modelled with a feedforward neural network. Transition function parameters $\kappa$ are trained with mean absolute distance loss against the compressed transitions from the environment. 

\paragraph{Reward network} $R_{o}: \mathcal{Z} \times \mathcal{A} \rightarrow \mathbb{R}$ maps state-action pairs to rewards. Similarly to the transition function, $R_{o}$ is trained with mean absolute distance against the real rewards from the environment.

\paragraph{Policy network} $\pi_{\theta}: \mathcal{Z} \rightarrow \mathcal{P}(\mathcal{A} | \mathcal{Z})$ outputs a discrete probability distribution over all actions in a given state, given the compressed latent representation of that state. 

\paragraph{Value network} $V_{\phi}: \mathcal{Z} \rightarrow \mathbb{R}$ learns state values under policy $\pi_{\theta}$. Similarly to actor-critic algorithms, the value network can be supervised with a variety of techniques. For simplicity, we choose MC policy rollout values (ie. rewards gathered during the previous episode). 

\paragraph{Memory modules} Behaviour memory stores only the most recent trajectories and is used as a basis for learning the policy and value approximators. Similarly to an experience buffer \cite{lin1992self}, WM memory stores some fixed amount of recent transitions, which do not have to be distributed according to the policy. Those policy-agnostic transitions are used to train the WM components of the agent.

\begin{figure}[h!]
    \centering
    \begin{algorithm}[H]
      \caption{WMPG episode
        \label{alg:tmAC}}
      \begin{algorithmic}[1]
        \Statex
    \Function{WMPG episode()}{}
    \parState{$s = env.reset()$ $\quad$ \% reset the environment} 
    \parWhile{$not\ terminal$}
    \parState{$z = Z_{\mu}(s)$ $\quad$ \% compress the state representation using $Z_{\mu}$} 
    \parState{$a \sim \pi_{\theta}(z)$ $\quad$ \% sample an action from the current policy}
    \parState{$s', r = env.step(a)$  $\quad$ \% execute the sampled action in the environment}
    \parState{$z' = Z_{\mu}(s')$ $\quad$ \% compress the future states before adding experience to the memory}
    \parState{$ \mathcal{D}_{\Pi} \leftarrow (z, a, r, z')$ $\quad$ \% add experience to the on-policy memory}
    \If{$\ len(\mathcal{D}_{\Pi}) > batch \ size$}
    \parState{$ \mathcal{D}_{WM} \leftarrow \mathcal{D}_{\Pi}$  $\quad$ \% add the on-policy data to the WM memory}
    \parWhile{$\ learning \ iterations$}
    \parState{$\dot{z}, \dot{a}, \dot{r}, \dot{z}^{'} \sim \mathcal{D}_{WM}$ $\quad$ \% sample transitions for WM training}
    \parState{$\kappa \leftarrow \kappa - \alpha \nabla_{\kappa}|T_{\kappa}(\dot{z},\dot{a}) - \dot{z}^{'}|$ $\quad$ \% train the transition network}
    \parState{$o \leftarrow o - \alpha \nabla_{o}|R_{o}(\dot{z},\dot{a}) - \dot{r}|$ $\quad$ \% train reward the network}
    \parState{$\phi \leftarrow \phi - \alpha \nabla_{\phi}|V_{\phi}(z) - V(\mathcal{D}_{\Pi})|$ $\quad$ \% train the value network}
    \parState{$\theta \leftarrow \theta + \alpha \nabla_{\theta}\hat{J}(\theta, z)$ $\quad$ \% train the policy using steps described in Section \ref{pgWM}}
    \EndparWhile
    \parState{$\mathcal{D}_{\Pi} = [ \ ]$ $\quad$ \% wipe the on-policy memory}
    \EndIf
    \EndparWhile
    \EndFunction
    \end{algorithmic}
    \end{algorithm}
\caption{WMPG learning episode. Unrolling the policy in imagination allows for multiple parameter updates using one batch of the on-policy data $\mathcal{D}_{\Pi}$. For detailed implementation, we point the reader to: \url{https://github.com/WMPG-paper/WMPG}.}
\label{algo}
\end{figure}

\section{Gradient Estimation in WMPG}
\label{pgWM}

WMPG approximates $\nabla_{\theta}J(\theta, s)$ using $k$ actions sampled without replacement from the world model. For any $k \leq |\mathcal{A}|$,  $\nabla_{\theta}\hat{J}(\theta, s)$ can be calculated by without-replacement MC estimator (\cite{horvitz1952generalization}; \cite{schulman2015trust}; \cite{shah2018without}; \cite{kool2019buy}):

\begin{equation}
\label{eqn:ffff}
    \nabla_{\theta}\hat{J}(\theta,s) = \sum_{i=1}^k \frac{\pi_{\theta}(a_i|s)}{ \Omega(a_i|\pi_{\theta},k,s)} Q^{\pi_{\theta}}(s,a_i) \nabla_{\theta} \log \pi_{\theta} (a_i|s)
\end{equation}

\begin{figure}[h!]
    \centering 
\begin{subfigure}{0.325\textwidth}
  \includegraphics[width=0.935\linewidth]{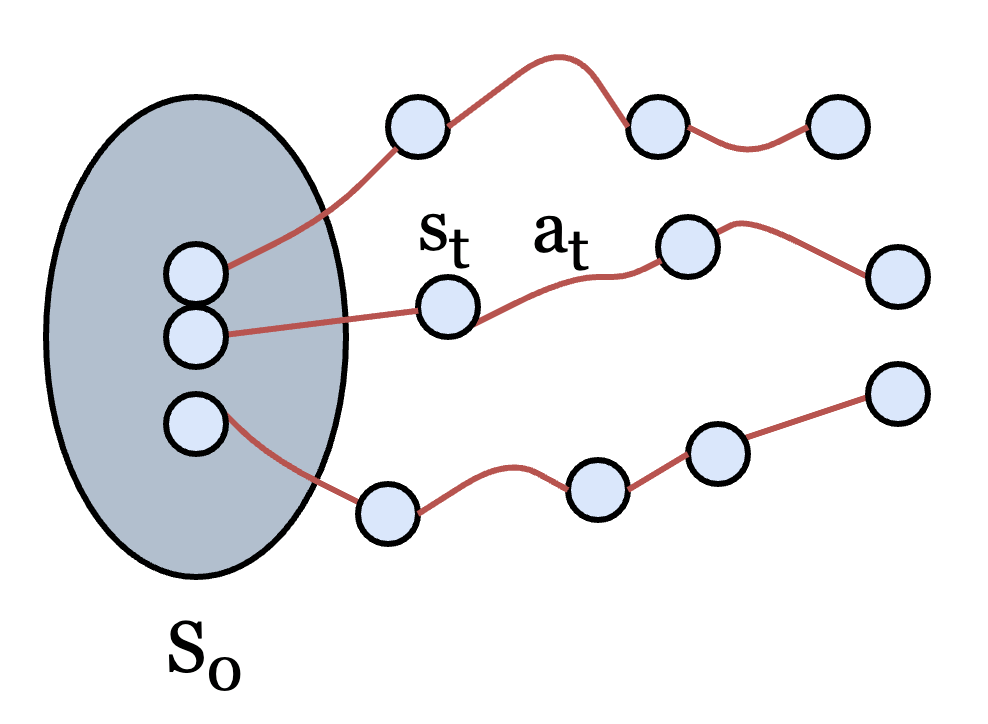}
  \caption{}
  \label{fig:horizon11}
\end{subfigure}
\begin{subfigure}{0.325\textwidth}
  \includegraphics[width=0.935\linewidth]{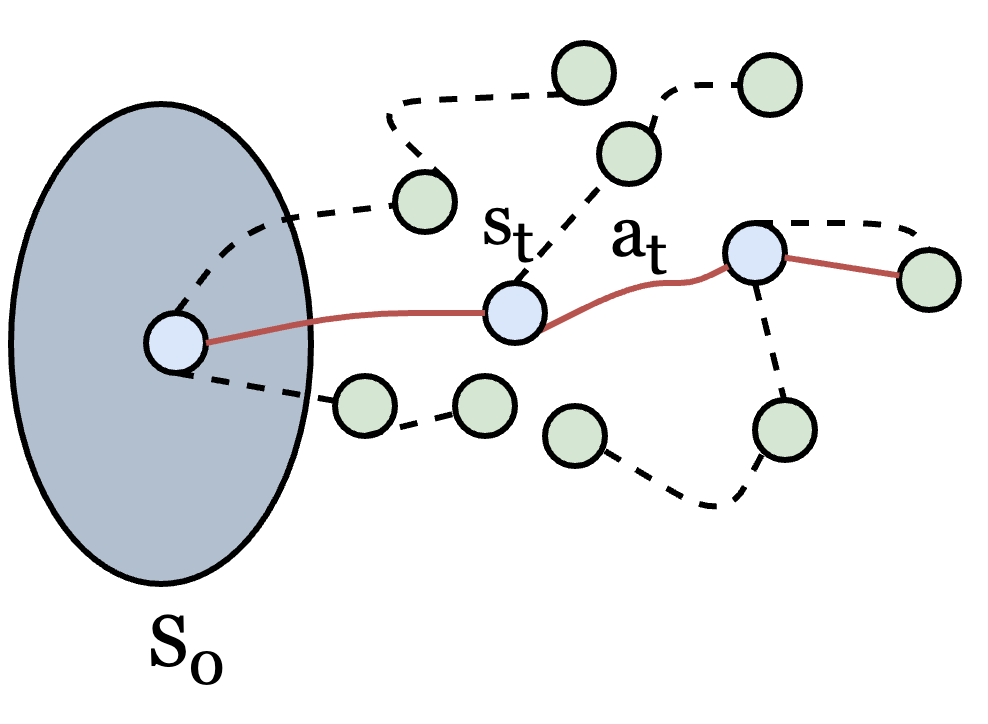}
  \caption{}
  \label{fig:horizon12}
\end{subfigure}
\begin{subfigure}{0.33\textwidth}
  \includegraphics[width=\linewidth]{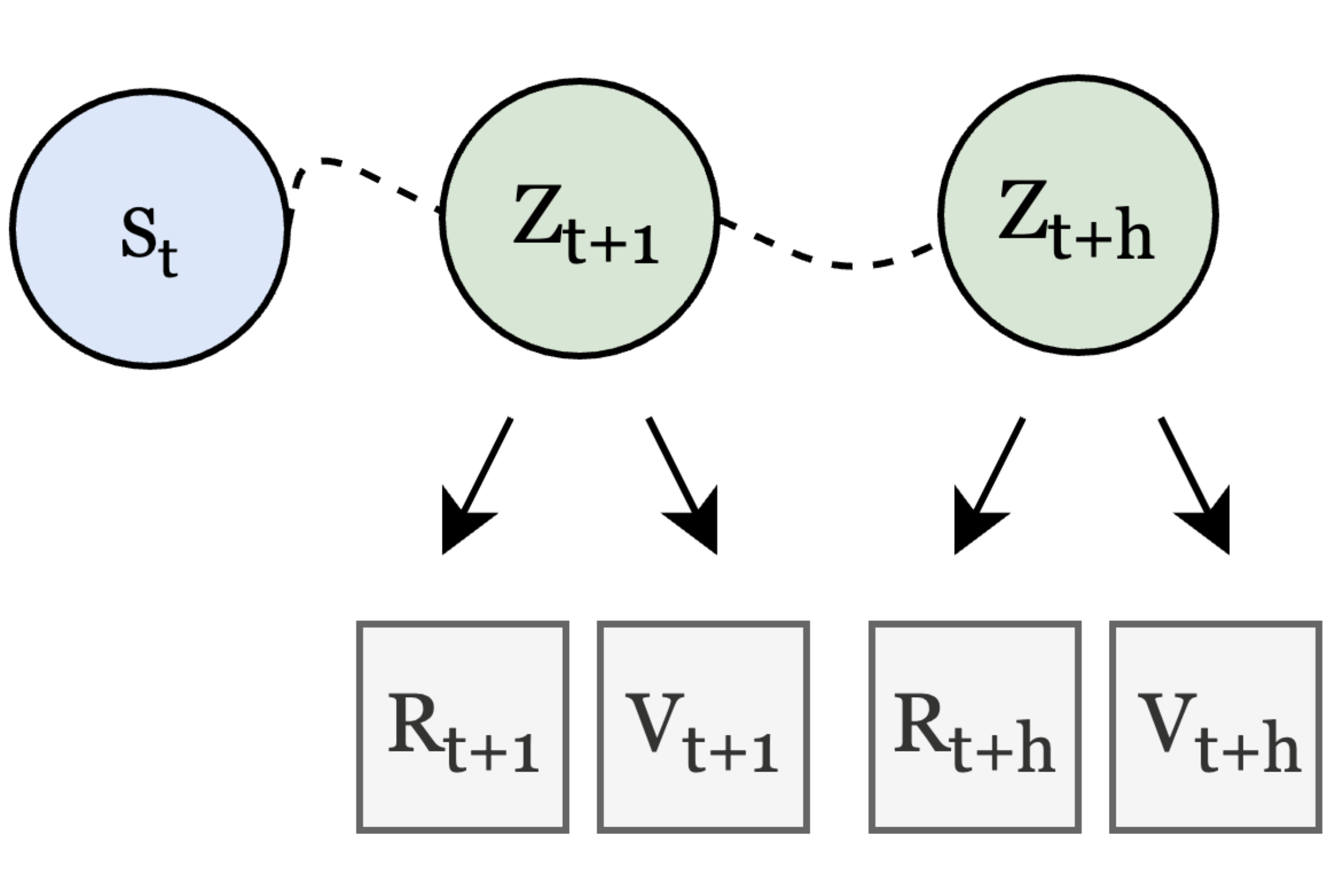}
  \caption{}
  \label{fig:horizon13}
\end{subfigure}
\caption{WMPG combines acting in the environment with learning in the imagination. a) Trajectories are sampled from the environment (represented by red lines) and gathered into a batch. b) Agent decides how many without-replacement actions should be sampled per batched state and imagines a fixed-length trajectories for each sampled action. c) Q-values of the sampled actions are approximated using TD($\lambda$).}
\label{fig:images1234}
\end{figure}

\noindent Where $i$ indicates the concurrent actions sampled without replacement from the WM and $\Omega(a_i|\pi_{\theta},k,s)$ is the inclusion probability of action $a_i$ (ie. probability that action $a_i$ was sampled without replacement given $k$ samples and policy $\pi_{\theta}$ in state $s$). For any fixed $k$ the estimator is an unbiased approximator of the exact expectation, which we show in Appendix \ref{HT}. Conveniently, for $k=1$ the estimator becomes equal to a vanilla MC estimator, while for $k = |A|$ the estimator becomes equal to the expected value over the sampled domain. Figure \ref{fig:images1234} shows how the WM is used to calculate the without-replacement approximator.

Intuitively, the without-replacement estimator allows for efficient evaluation of the expectation by not oversampling the high-probability domain. If the policy has low entropy, then sampling with replacement can yield repeated results, which in the context of policy gradient is uninformative.

Formally, the estimator reduces the approximation variance by remodelling the policy gradient from the sum of independent random variables to the weighted sum of dependent random variables. This slows the diminishing of the variance reduction associated with drawing additional samples. For details, we refer the reader to Appendix \ref{VAR}. 

In the following paragraphs, we detail how the policy gradient is approximated in WMPG. We describe the process in four steps: 1) \textit{choosing a $k$}; 2) \textit{Q-value approximation}; 3) \textit{baseline calculation}; and 4) \textit{weight normalization}. 

\paragraph{Choosing $k$} The without-replacement estimator is unbiased for any fixed size of $k$. Since $\nabla_{\theta}J(\theta,s)$ is approximated for every batched state independently, $k$ can be chosen per evaluated state. Thus, it seems natural to ask: is there a way of choosing a $k$ that well-balances the robustness of the estimator with the required compute? While sampling theory offers frameworks to tackle such questions (\cite{hesterberg1988advances}; \cite{shah2018without}), we consider the following simple heuristics: constant $k$ throughout the training; $k$ decreasing throughout the training; and $k$ dependent on the entropy of policy at given state.

\paragraph{Q-value approximation} WMPG uses the transition, reward and value networks to approximate the unknown Q-values. For each sampled action, WMPG imagines a trajectory of length $h$ and estimates the respective Q-value with TD($\lambda$):

\begin{equation}
    \hat{Q}^{\pi_{\theta}}(s,a, \lambda, h) = (1 - \lambda) \sum_{n=1}^{h-1} \lambda^{n-1} TD(n|s,a) + \lambda^{N-1} TD(h|s,a)
\end{equation}

Where $TD(n|s,a)$ denotes $n$-step temporal difference value, given the starting state-action pair and following policy $\pi_{\theta}$.

\paragraph{Baseline Variance Reduction} WMPG uses the without-replacement value approximation as a baseline for the policy gradient:

\begin{equation}
    \hat{V}(s) = \sum_{i=1}^k \frac{\pi_{\theta}(a_i|s)}{ \Omega(a_i|\pi_{\theta},k,s)}\hat{Q}^{\pi_{\theta}}(s,a_i, \lambda, h)
\end{equation}

Since the importance weights depend on the sampled actions, the resulting estimator becomes biased. Such bias can be corrected by introducing an action-dependent term $C(s,a_i) = (1 + \frac{\pi_{\theta}(a_i|s)}{ \Omega(a_i|\pi_{\theta},k,s)} - \pi_{\theta}(a_i|s))$ \cite{kool2019buy}:
\begin{equation}
        \nabla_{\theta}\hat{J}(\theta,s) =
        \sum_{i=1}^k \frac{\pi_{\theta}(a_i|s)}{ \Omega(a_i|\pi_{\theta},k,s)} \left(C(s,a_i) \hat{Q}^{\pi_{\theta}}(s,a_i, \lambda, h) - \hat{V}(s) \right) \nabla_{\theta} \log \pi_{\theta}(a_i|s)
\label{eqn:ffff12222}
\end{equation}

When all actions are sampled, then the baseline becomes independent of the sampling process. As such, if $k = |\mathcal{A}|$ then $\frac{\pi_{\theta}(a_i|s)}{ \Omega(a_i|\pi_{\theta},k,s)} = \pi_{\theta}(a_i|s)$, and in consequence $C(s,a_i) = 1$ for all $i$.

\paragraph{Weight normalization}

The weighted probabilities do not generally sum to one. While such calculation leads to unbiased estimation, it has high variance. As the policy gradient is typically evaluated once before the parameter update, we are interested in low-variance approximation of $\nabla_{\theta}\hat{J}(\theta,s)$. The without-replacement expectation estimator can be normalized by the sum of the sampled weights \cite{kool2019stochastic}. To this end, we incorporate the normalization into the state policy gradient calculations according to:

\begin{equation}
\label{eqn:pg1}
    \nabla_{\theta}\hat{J}(\theta,s) = \sum_{i=1}^k \frac{\pi_{\theta}(a_i|s)}{ \Omega(a_i|\pi_{\theta},k,s)} \left( \frac{ \hat{Q}^{\pi_{\theta}}(s,a, \lambda, h)}{W_i(s)} - \frac{\hat{V}(s)}{W(s)}  \right) \nabla_{\theta} \log \pi_{\theta}(a_i|s)
\end{equation}

\noindent With $W(s) = \sum_{i=1}^k \frac{\pi_{\theta}(a_i|s)}{ \Omega(a_i|\pi_{\theta},k,s)}$ and $W_i(s) = W(s) -  \frac{\pi_{\theta}(a_i|s)}{ \Omega(a_i|\pi_{\theta},k,s)} + \pi_{\theta}(a_i|s)$. While the normalization introduces bias, the estimator remains consistent \cite{hesterberg1988advances} and is reported to perform better in the context of REINFORCE \cite{kool2019stochastic}.

\section{Experiments}
\label{sec:experiments}

We conducted experiments to test sample efficiency (ie. number of the required environment episodes to achieve a certain performance) of the proposed approach as compared to standard policy gradient algorithms. We test the WMPG sample efficiency in two contexts: 1. \textit{compact state representation}; and 2. \textit{high-dimensional state representation}. To limit the scope of the hyperparameter search, we do not search for WMPG hyperparameters that are available for AC. For those parameters, WMPG uses the values found for the AC implementation. The results are presented on Figure \ref{fig:images12341}; for further details regarding the experimental setting, see Appendix \ref{EXP}.

\begin{figure}[h!]
    \centering 
\begin{subfigure}{0.32\textwidth}
  \includegraphics[width=\linewidth]{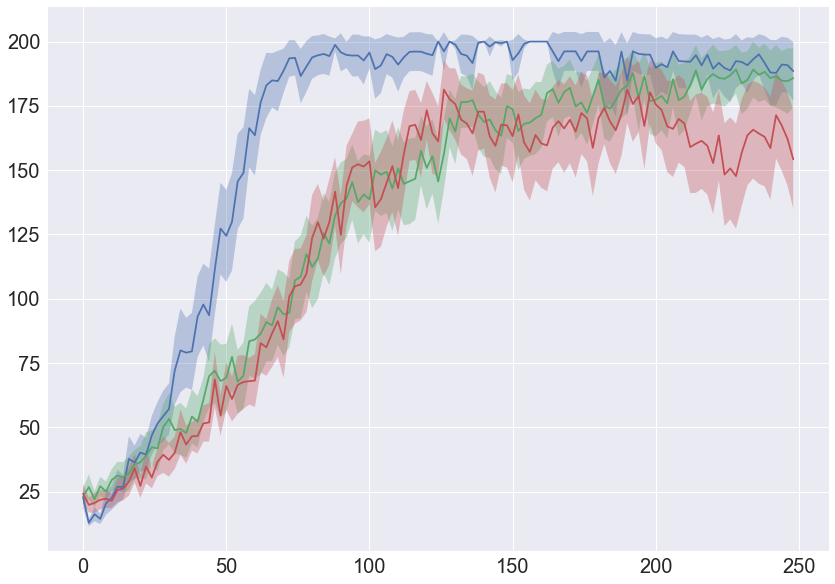}
  \caption{CartPole; 50 seeds}
  \label{fig:learning_curve11}
\end{subfigure}
\begin{subfigure}{0.32\textwidth}
  \includegraphics[width=\linewidth]{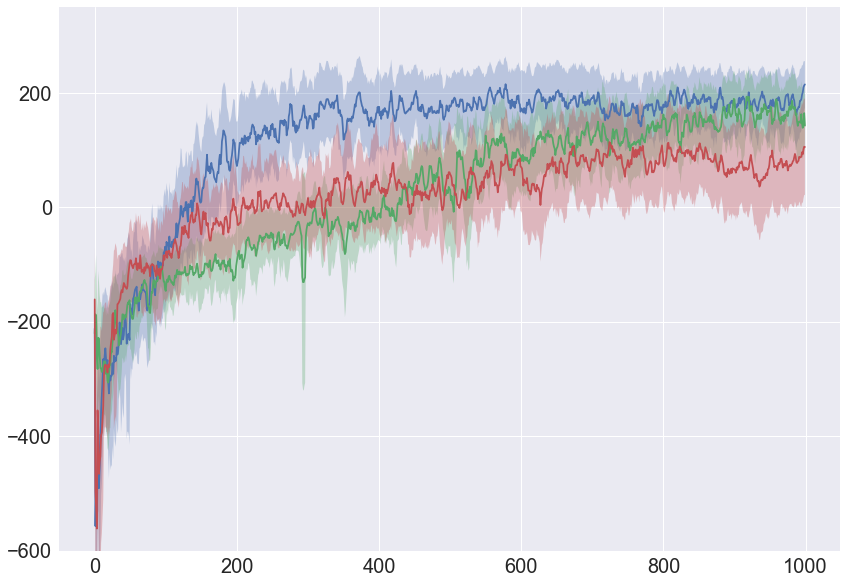}
  \caption{LunarLander; 10 seeds}
  \label{fig:learning_curve12}
\end{subfigure}
\begin{subfigure}{0.32\textwidth}
  \includegraphics[width=\linewidth]{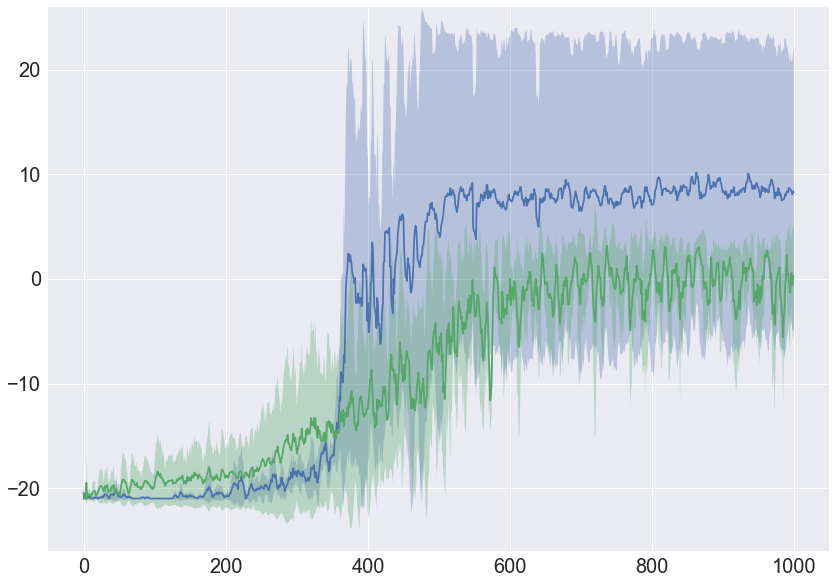}
  \caption{Pong; 2 seeds}
  \label{fig:learning_curve13}
\end{subfigure}
\caption{Learning curves. $y$-axis denotes the average accumulated reward and $x$-axis denotes the environment episode number. [Blue] is WMPG; [green] is AC; and [red] is MAC. Shadow denotes the distance of two standard deviations of the average. a) and b) Given compact value preserving state representation, WMPG yields robust performance gain. c) Encoding the state representation with a VAE can result in faster policy convergence as compared to learning on the original problem representation (\cite{ha2018recurrent}; \cite{hafner2019learning}). In our experiments, only one seed converged to the optimal policy of 21 points per episode. Thus, we hypothesize that learning control via VAE compressed representations might be more prone to local optima with suboptimal policies.}
\label{fig:images12341}
\end{figure}

\paragraph{Compact state representation} We treat the problem of representation learning in WM as orthogonal to this paper. As such, we test WMPG in environments where representation learning can be omitted (ie. state embedding function can be represented by an identity mapping) and the state representation is guaranteed to be value preserving. By doing so, we look for the sample efficiency gains which are directly attributable to mechanisms of WMPG, that is without-replacement approximation of $\nabla_{\theta}J(\theta,s)$ and TD($\lambda$) approximation of Q-values using online-learned transition and reward mappings. For the compact state representation testing we choose the CartPole and LunarLander environments, run for respectively 250 and 1000 environment episodes.

\paragraph{High-dimensional state representation} Further, we test WMPG against AC on high-dimensional visual input of Arcade Learning Environment (ALE) Pong \cite{bellemare2013arcade} given 1000 environment episodes. Here, we learn the representations with a variational reconstruction loss placed in the pixel space \cite{kingma2013auto}. The VAE is pre-trained using data generated over 10 episodes following a random policy. Learning on a compressed latent representation effectively decreases the size of policy search problem and is expected to yield efficiency gains independently of other mechanisms used. Therefore, to limit the impact that state embedding has on the results, we train AC on the VAE state representations used for WMPG.

\paragraph{Ablations}

\begin{figure}[h!]
    \centering 
\begin{subfigure}{0.32\textwidth}
  \includegraphics[width=\linewidth]{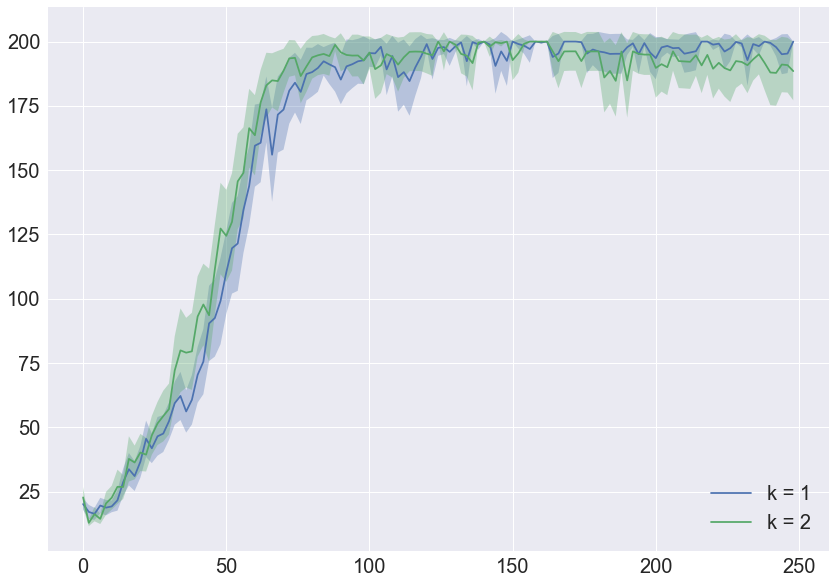}
  \caption{CartPole; 50 seeds}
  \label{fig:horizon11}
\end{subfigure}
\begin{subfigure}{0.32\textwidth}
  \includegraphics[width=\linewidth]{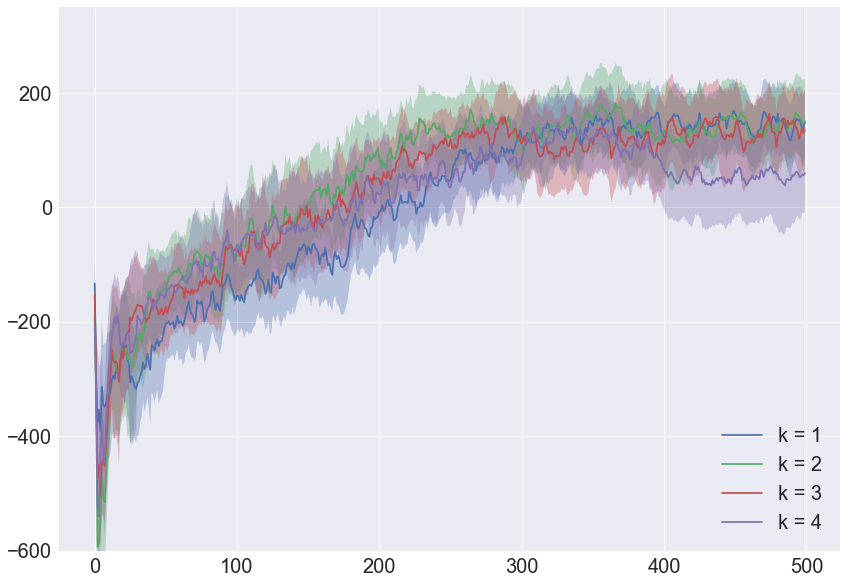}
  \caption{LunarLander; 10 seeds}
  \label{fig:horizon12}
\end{subfigure}
\begin{subfigure}{0.32\textwidth}
  \includegraphics[width=\linewidth]{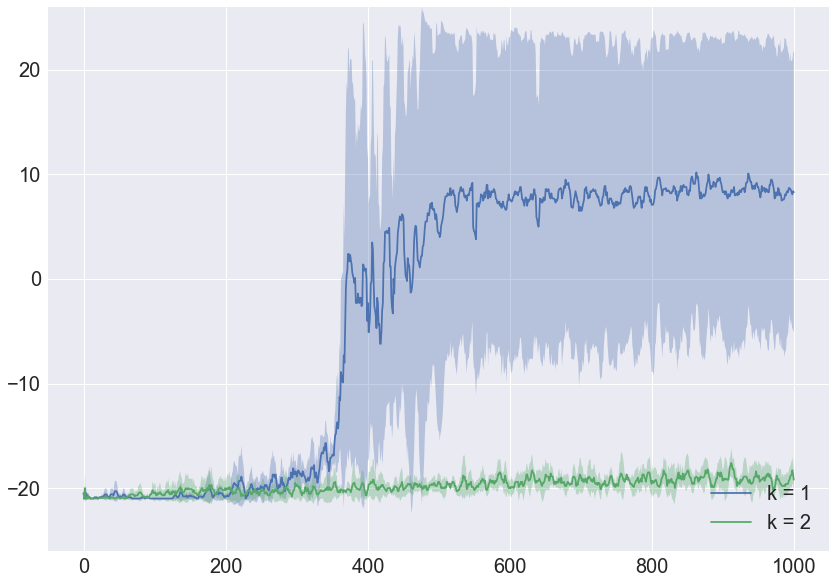}
  \caption{Pong; 2 seeds}
  \label{fig:horizon13}
\end{subfigure}
\caption{WMPG performance for different number of without-replacement actions. $y$-axis denotes the average accumulated reward and $x$-axis denotes the environment episode number. [Blue] is $k = 1$; [green] is $k = 2$; [red] is $k = 3$; and [purple] is $k = 4$. If only one action is sampled, then we use the value network at given state as the baseline. Shadow denotes the distance of two standard deviations of the average. a) and b) Q-values of the imagined actions are biased due to TD($\lambda$) and WM components. Therefore, imagining additional actions with a WM might induce a bias-variance trade-off, especially in the early stages of training. c) Imagining all actions yielded almost random results on all tested Pong seeds. We explain this with the fact that WMPG nullifies the gradient when Q-values of different action are equal (look Equation \ref{eqn:ffff12222}). Since Pong has relatively sparse rewards, it might be that short-horizon TD($\lambda$) approximations yield similar Q-values for many actions at a given state.}
\label{fig:diffk}
\end{figure}

\begin{figure}[h!]
    \centering 
\begin{subfigure}{0.32\textwidth}
  \includegraphics[width=\linewidth]{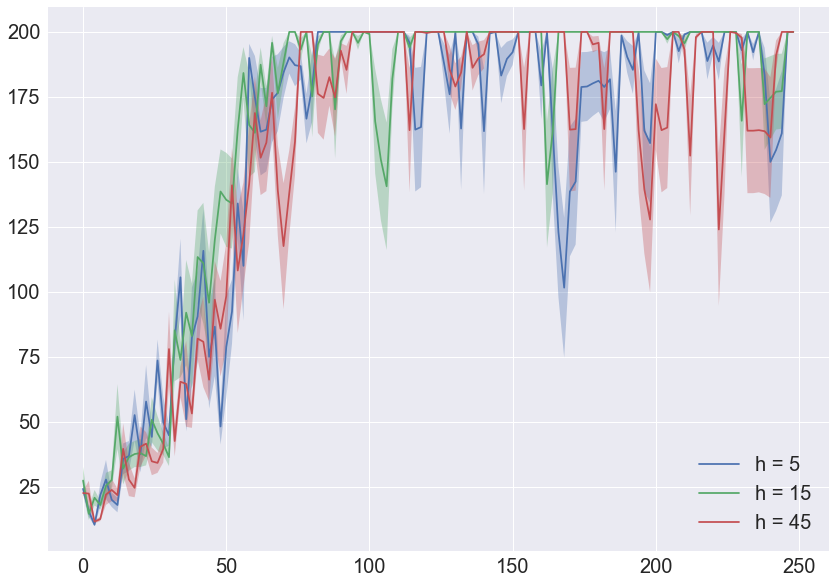}
  \caption{$\lambda=0.25$}
  \label{fig:horizon12}
\end{subfigure}
\begin{subfigure}{0.32\textwidth}
  \includegraphics[width=\linewidth]{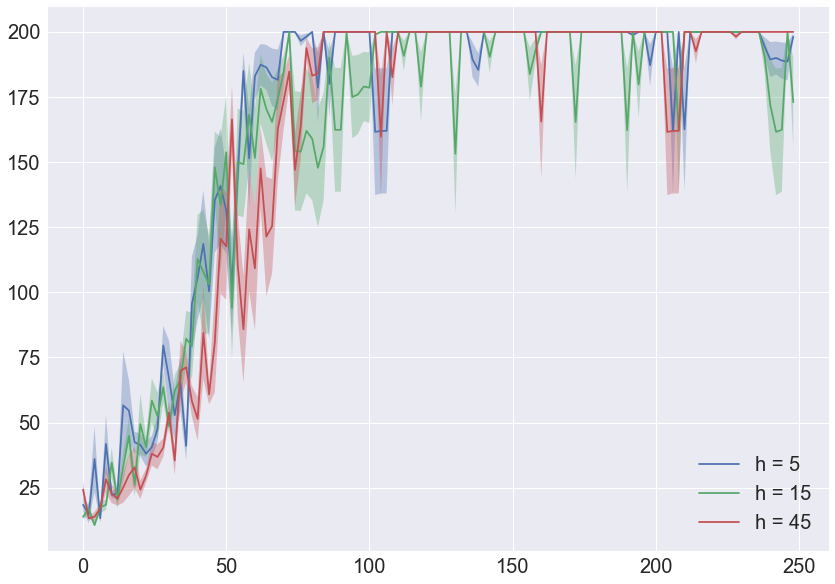}
  \caption{$\lambda=0.5$}
  \label{fig:horizon12}
\end{subfigure}
\begin{subfigure}{0.32\textwidth}
  \includegraphics[width=\linewidth]{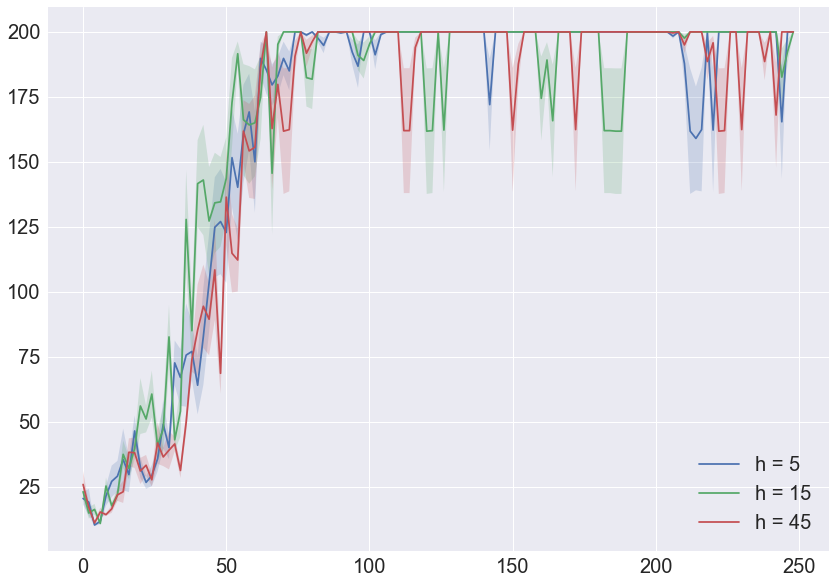}
  \caption{$\lambda=0.75$}
  \label{fig:horizon12}
\end{subfigure}
\medskip
\begin{subfigure}{0.32\textwidth}
  \includegraphics[width=\linewidth]{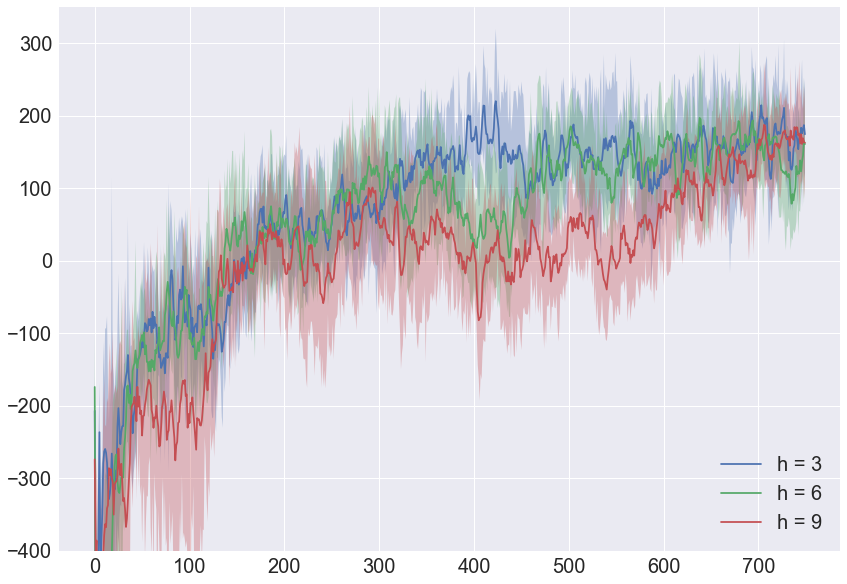}
  \caption{$\lambda=0.25$}
  \label{fig:horizon11}
\end{subfigure}
\begin{subfigure}{0.32\textwidth}
  \includegraphics[width=\linewidth]{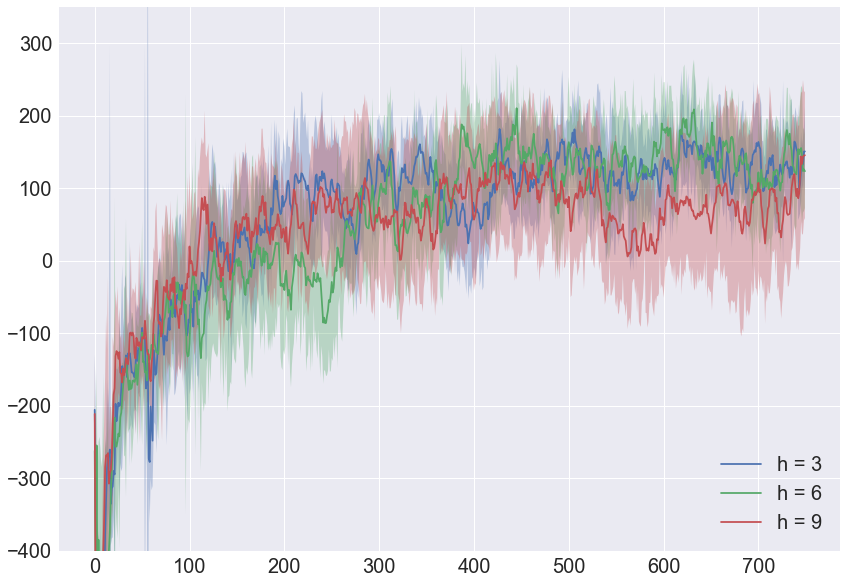}
  \caption{$\lambda=0.5$}
  \label{fig:horizon11}
\end{subfigure}
\begin{subfigure}{0.32\textwidth}
  \includegraphics[width=\linewidth]{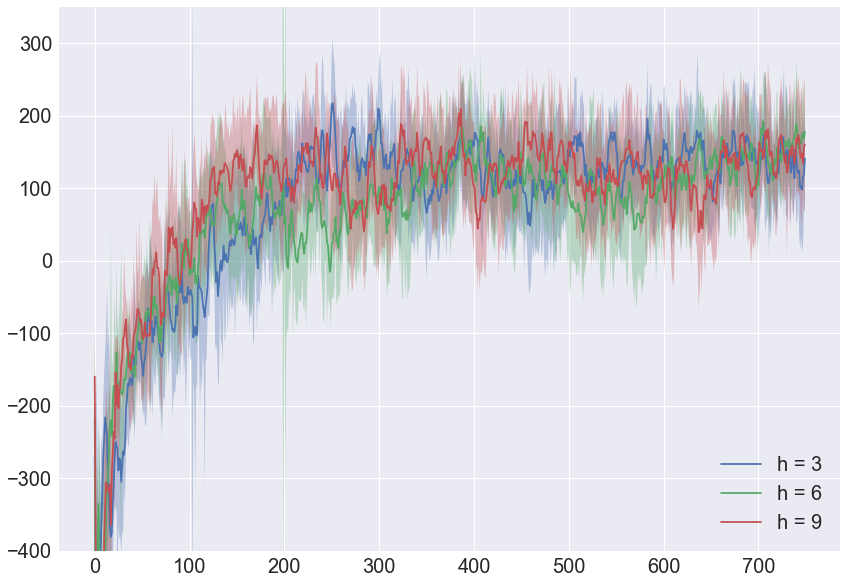}
  \caption{$\lambda=0.75$}
  \label{fig:horizon11}
\end{subfigure}
\caption[Baseline comparison on CartPole]{WMPG learning curves given different horizon lengths and $\lambda$. $y$-axis denotes the average reward accumulated over 8 seeds and $x$-axis denotes the episode number. Colours denote different imagination horizon lengths. Shadow denotes the distance of two standard deviations of the average. a), b), c) depict CartPole. There, [blue]: $h = 5$; [green]: $h = 15$; and [red]: $h = 45$. d), e) and f) depict LunarLander. There, [blue]: $h = 3$; [green]: $h = 6$; and [red]: $h = 9$.}
\label{fig:images997}
\end{figure}

WMPG integrates several mechanisms that could potentially lead to better sample efficiency. To gain a better understanding of the contribution of each mechanism towards the final performance, we performed additional experiments. Firstly, we look how sampling additional actions without replacement effects the WMPG performance (Figure \ref{fig:diffk}). Furthermore, we investigate the learning curves of WMPG given $k = |\mathcal{A}|$ and different values of $h$ and $\lambda$ (Figure \ref{fig:images997}).

\section{Related Work}
\label{sec:related_work}

\paragraph{World models}

VPN simulates multi-step returns to approximate and backup the Q-values over a horizon \cite{oh2017value}. Ha et al. \cite{ha2018recurrent} uses evolutionary approaches to search for policy in the compact WM representation of the environment. Similarly, PLANET \cite{hafner2019learning} uses non-differentiable optimization to plan within the learned WM. Value iteration was successfully applied to learned WM representations \cite{van2020plannable}. MuZero \cite{schrittwieser2019mastering} leverages the learned WM to perform Monte-Carlo tree search in the latent representation. SimPLe \cite{kaiser2019model} and Dreamer \cite{hafner2019dream} calculate the policy gradient with trajectories imagined by the WM. 

\paragraph{Policy gradients}

A3C leverages parallel agents to gather many trajectories from the environment, but calculates the policy gradient with MC \cite{mnih2016asynchronous}. TRPO vine procedure \cite{schulman2015trust} and \cite{kool2019stochastic} consider sampling actions without replacement at sampled states. There, agent is allowed to 'rewind' back to a given state. Kool et al. propose to use the without-replacement samples to calculate an informed baseline \cite{kool2019buy}. MAC uses Q-network approximations to calculate $\nabla_{\theta}J(\theta,s)$ exactly (ie. iterating over all actions) \cite{asadi2017mean}. TRPO \cite{schulman2015trust} updates policy parameters with natural gradient \cite{kakade2002natural}, but was found to be costly for bigger models. ACKTR \cite{wu2017scalable} calculates the trust-region update using less costly Kronecker-factored approximation. 

\section{Conclusion}
\label{sec:conclusion}

Imagining trajectories with latent models allows for gradient approximation strategies unavailable to traditional agents. In particular, we showed that WM can be used to sample and evaluate many actions per analyzed state. Contrary to MC, the without-replacement expectation estimator does not redundantly resample the-high probability domain. Our results indicate that WM-based without-replacement trajectory sampling is a promising approach for better sample efficiency in gradient-based policy search.

Ablation studies have revealed a surprisingly nonlinear relation between the number of actions sampled without replacement and the agent performance in the context of WM-based learning. Imagining trajectories indeed reduces variance of policy gradient, but at the cost of bias that is inherent to WM approximations. Thus, we suspect that increasing the number of sampled action creates a bias-variance trade-off, with bias magnified by each imagined trajectory. Furthermore, our results indicate that varying the number of imagined samples yields better performance than any constant number of samples. We believe that applying recent advances in incremental without-replacement sampling (\cite{shah2018without}; \cite{shi2020incremental}) to gradient-based policy search might be a promising research direction. 

WMPG is an approach for sample efficient reinforcement learning using world models. The approach learns a world model which is used to imagine trajectories sampled without replacement. These are used to estimate policy gradients with a low variance without-replacement expectation estimator. We showed that the proposed approach can yield increased sample efficiency as compared to AC and MAC. 

\section*{Broader Impact}

This paper considers the problem of variance reduction in policy gradient approximation. As such, authors believe that there are no societal consequences specific to this work.


\small

\bibliography{bib}

\appendix

\newpage

\section{Unbiasedness of the without-replacement expectation estimator}
\label{HT}

We assume $X$ to be a discrete random variable with domain of $|\mathcal{A}|$ values ($x_1, ..., x_{|\mathcal{A}|}$) and respective probabilities ($p_1, ..., p_{|\mathcal{A}|}$). The expected value of $X$ is given by:

\begin{equation}
\label{ht0}
    \mathbb{E}\left[ X \right] = \sum_{a=1}^{\mathcal{|A|}} p_a x_a
\end{equation}

We denote $\hat{x}$ as the following estimator of $\mathbb{E}\left[ X \right]$:

\begin{equation}
    \hat{x} = \sum_{i = 1}^{k} \beta_{i} x_i
\end{equation}

Where $k$ is the number of without-replacement samples drawn from $X$, $x_i$ is the $i^{th}$ sample drawn without replacement and $\beta_{i}$ is a weight independent of determined by $x_i$. We want to show that for any constant value of $k\leq |\mathcal{A}|$, $\hat{X}$ is unbiased is expectation.

\begin{equation}
\label{ht1}
    \mathbb{E}\left[ \hat{X} \right] = \mathbb{E}\left[ X \right]
\end{equation}

We first note that $\hat{X}$ is defined as $k$ without-replacement samples from $X$. Therefore, $\hat{X}$ has a domain of $\mathcal{A}\choose{k}$ $= K$ values denoted as $\hat{X} = (\hat{x}_{1}, ..., \hat{x}_{K})$ and respective probabilities $(q_1, ..., q_{K})$. Therefore:

\begin{equation}
    \mathbb{E}\left[ \hat{X} \right] = \sum_{j=1}^{K} q_j \hat{x}_{j} = \sum_{j=1}^{K} q_j \sum_{i = 1}^{k} \beta_{i} x_i = \sum_{j=1}^{K} q_j \beta_{1} x_{1}^{j} + q_j \beta_{2} x_{2}^{j} + ... + q_j \beta_{k} x_{k}^{j}
\end{equation}

Now, we note that $x_a$ can be sampled in some number of $\hat{x}$. We denote the set of indices of $\hat{X}$ that contain $x_a$ as $L$. Then, by expanding the sum:

\begin{equation}
\label{ht2}
    \mathbb{E}\left[ \hat{X} \right] = \sum_{a=1}^{\mathcal{|A|}} \beta_a x_a \sum_{j \in L} q_j
\end{equation}

The above operation explicitly assumes that $\beta_a$ is independent of $j$. Combining Equations \ref{ht0}, \ref{ht1} and \ref{ht2} yields:

\begin{equation}
    \sum_{a=1}^{\mathcal{|A|}} p_a x_a = \sum_{a=1}^{\mathcal{|A|}} \beta_a x_a \sum_{j \in L} q_j
\end{equation}

Thus, when $\beta_a = \frac{p_a}{\sum_{j \in L} q_j}$, the estimator is an unbiased estimator of $\mathbb{E}\left[ X \right]$.

\section{Variance of the without-replacement policy gradient estimator}
\label{VAR}

Here, we show why calculating policy gradient without replacement might lead to a better variance reduction stemming from increase of samples. To not overload the notation, we assume the Q-values to be known.

Given a batch of states $D_s = (s_1, s_2, ..., s_{|D_s|})$, gradient $\nabla_{\theta}\hat{J}(\theta)$ is calculated with batch average. Without assuming independence between sampled states, variance of the policy gradient estimator is equal to:

\begin{equation}
    Var \left[  \nabla_{\theta}\hat{J}(\theta)\right] =  \frac{1}{|D_s|^{2}} \sum_{s \in D_s} \left( Var \left[ \nabla_{\theta}\hat{J}(\theta, s)\right] + \sum_{s' \neq s \in D_s} Cov \left[\nabla_{\theta}\hat{J}(\theta, s), \nabla_{\theta}\hat{J}(\theta, s')  \right] \right) 
\end{equation}

Where $\nabla_{\theta}\hat{J}(\theta, s)$ denotes evaluation of the policy gradient at state $s$. If $\nabla_{\theta}\hat{J}(\theta, s)$ is calculated via MC with $|D_a|$ samples, then:

\begin{equation}
\label{bvvv}
    Var \left[  \nabla_{\theta}\hat{J}^{mc}(\theta, s)\right] = \frac{Var \left[  Q^{\pi_{\theta}}(s,a) \nabla_{\theta} \log \pi_{\theta} (a|s) \right] }{|D_a|}
\end{equation}

Taking derivative wrt. $|D_a|$ reveals that the decrease of variance stemming from taking more samples is rapidly diminishing. Furthermore, if the variance is non-zero, then $Var \left[  \nabla_{\theta}\hat{J}^{mc}(\theta, s)\right]$ converges to 0 for $|D_a| = \infty$. 

If we treat $|D_a|$ as the number of actions sampled without replacement, then for $|D_a| = |\mathcal{A}|$ (ie. when sampling all actions) $\nabla_{\theta}\hat{J}(\theta, s)$ is calculated with exact expectations:

\begin{equation}
\label{rep123}
    \nabla_{\theta}\hat{J}^{wr}(\theta, s) = \sum_{a \in |\mathcal{A}|} \nabla_{\theta} \pi_{\theta}(a|s) Q^{\pi_{\theta}}(s, a)
\end{equation}

Thus, given $|D_a| = |\mathcal{A}|$ and known Q-values, it follows that $Var \left[  \nabla_{\theta}\hat{J}^{wr}(\theta, s)\right] = 0$. 
As such, assuming $|D_a| < |\mathcal{A}|$ it might be the case that $Var \left[  \nabla_{\theta}\hat{J}^{wr}(\theta, s)\right] < Var \left[  \nabla_{\theta}\hat{J}^{mc}(\theta, s)\right]$. Borrowing the notation from Appendix \ref{HT}, the variance of without-replacement estimator is equal to:

\begin{equation}
    Var \left[ \nabla_{\theta}\hat{J}^{wr}(\theta, s) \right] = \sum_{j=1}^{K} q_j \left( \sum_{i = 1}^{|D_a|} \beta_{i} x_i \right)^2 - \left( \nabla_{\theta}J(\theta, s) \right)^2
\end{equation}

For further reading on the without-replacement estimation variance, we point the reader to (\cite{horvitz1952generalization}; \cite{kool2019stochastic}; \cite{shah2018without}).

\section{Experiment details}
\label{EXP}

Below, we detail the implementation of the experiments. Table \ref{n_params} lists the searched hyperparameters for each model:

\begin{table}[H]
\centering
 \begin{tabular}{||c c c c c||} 
 \hline
  Name & Description & AC & MAC & WMPG \\
  $\pi_{size}$ & Policy network size & x & x & x \\
  $V_{size}$ & Value network size & x & (x) & x \\
  $T_{size}$ & Transition network size &  &  & (x) \\
  $R_{size}$ & Reward network size & & & (x) \\
  $O_{\pi}$ & Optimizer for $\pi$ & (x) & (x) & x \\
  $O_{V}$ & Optimizer for value & (x) & (x) & x \\
  $O_{T}$ & Optimizer for transition &  &  & (x) \\
  $O_{R}$ & Optimizer for reward &  &  & (x) \\
  $\alpha_{\pi}$ & Learning rate for $\pi$ & (x) & (x) & x \\
  $\alpha_{V}$ & Learning rate for value & (x) & (x) & x \\
  $\alpha_{T}$ & Learning rate for transition &  &  & (x) \\
  $\alpha_{R}$ & Learning rate for reward &  &  & (x) \\
  $I_{G}$ & General learning iterations &   & (x) & (x) \\
  $I_{V}$ & Value learning iterations &  & (x) & (x) \\
  $I_{G}$ & WM learning iterations &  &  & (x) \\
  $h$ & Horizon length &  &  & (x) \\
  $k$ & Sampled actions &  &  & (x) \\
  $\lambda$ & TD($\lambda$) controller &  &  & (x) \\
 \hline \hline
\multirow{3}{*}{\# of configs} & CartPole & 81 & 126 & 126 \\
& LunarLander & 9 & 24 & 36 \\
& Pong & 8 & NA & 4 \\
\hline
\end{tabular}
\caption[Hyperparameter tuning]{Number of tunable hyperparameters; x denotes that parameter is tunable; (x) denotes that parameter was tuned for experiments in this paper}
\label{n_params}
\end{table}

Both AC and MAC are implemented using two separate feedforward neural networks. Both value and Q-value networks are trained using MC policy rollout. 

All models use a discount factor of 0.99 and batch sizes of (CartPole - 32; LunarLander - 64; and Pong - 512). All models run on LunarLander and Pong use a multiplicative learning rate annealing with a rate of 0.99 and a step sizes of 125 and 5 respectively. 

Pong environment is implemented with a frame-skip of four and non-sticky actions. Finally, the Pong frames are preprocessed: size of the frame is decreased to (80, 80, 1) and the ball is enlarged by the factor of two. 

For further details regarding the implementation, we point the reader to the provided GitHub repository. Finally, all Pong agents use an entropy coefficient of 0.01 \cite{haarnoja2018acquiring}.

\subsection{AC}

Table below lists the hyperparameter values used for AC:

\begin{table}[H]
\centering
 \begin{tabular}{||c | c c c c c c c||} 
 \hline
 env & $\pi_{size}$ & $V_{size}$ & $\alpha_{\pi}$ & $\alpha_{V}$ & $O_{\pi}$ & $O_{V}$ & $I_G$ \\
 \hline \hline
 CartPole & 32 & 64 & .0025 & .005 & RMS & RMS & 1 \\
 LunarLander & 64 & 64 & .0025 & .0025 & RMS & RMS & 1 \\
 Pong & 512 & 512 & .001 & .001 & RMS & RMS & 1 \\
 \hline
\end{tabular}
\caption[AC CartPole setting]{Best performing AC configurations.}
\label{maasdc_grid}
\end{table}

\subsection{MAC}

Table below lists the hyperparameter values used for MAC:

\begin{table}[H]
\centering
 \begin{tabular}{||c | c c c c c c c c||} 
 \hline
 env & $\pi_{size}$ & $Q_{size}$ & $\alpha_{\pi}$ & $\alpha_{Q}$ & $O_{\pi}$ & $O_{Q}$ & $I_G$ & $I_V$ \\
 \hline \hline
 CartPole & 32 & [64,64] & .00125 & .005 & RMS & RMS & 3 & 3  \\
 LunarLander & 64 & [64,64] & .0025 & .005 & RMS & RMS & 5 & 3 \\
 \hline
\end{tabular}
\caption[MAC CartPole setting]{Best performing MAC configurations.}
\label{mac_grid}
\end{table}

\subsection{WMPG}

Table below lists the hyperparameter values used for WMPG:

\begin{table}[H]
\centering
 \begin{tabular}{||c | cc c c c c c c c||} 
 \hline
  env & $T_{size}$/$R_{size}$ & $\alpha_{T}$/$\alpha_{R}$ & k & $O_{R}$/$O_{R}$ & $I_G$ & $I_V$ & $I_{WM}$ & $h$ & $\lambda$ \\
 \hline \hline
 CartPole & 64 & .005 & 2 & Adam & 5 & 3 & 5 & 15 & 0.75  \\
 LunarLander & 96 & .005 & Decreas. & Adam & 4 & 3 & 5 & 3 & 0.75  \\
 Pong & 1028 & .002 & 1 & Adam & 5 & 1 & 1 & 3 & 0.75  \\
 \hline
\end{tabular}
\caption[WMPG CartPole setting]{Best performing configurations of WMPG. Other settings are taken from respective AC implementations.}
\label{WMPG_parame}
\end{table}

\end{document}